\documentclass[conference]{IEEEtran}
\IEEEoverridecommandlockouts
\usepackage{cite}
\usepackage{amsmath,amssymb,amsfonts}
\usepackage{algorithmic}
\usepackage{graphicx}
\usepackage{textcomp}
\usepackage{xcolor}
\usepackage{comment}
\usepackage{hyperref}
\def\BibTeX{{\rm B\kern-.05em{\sc i\kern-.025em b}\kern-.08em
    T\kern-.1667em\lower.7ex\hbox{E}\kern-.125emX}}
\usepackage{dsfont}

\renewcommand{\baselinestretch}{0.99}
\begin{document}

\title{Koopman Ensembles for \\Probabilistic Time Series Forecasting\\
\thanks{\noindent{This work was supported by Agence Nationale de la Recherche under grant ANR-21-CE48-0005 LEMONADE.}}
}

\author{\IEEEauthorblockN{Anthony Frion}
\IEEEauthorblockA{
\textit{Lab-STICC, IMT Atlantique} \\
Brest, France \\
anthony.frion@imt-atlantique.fr}
\and
\IEEEauthorblockN{Lucas Drumetz}
\IEEEauthorblockA{
\textit{Lab-STICC, IMT Atlantique} \\
Brest, France \\
lucas.drumetz@imt-atlantique.fr}
\and
\IEEEauthorblockN{Guillaume Tochon}
\IEEEauthorblockA{
\textit{LRE EPITA} \\
 Le Kremlin-Bicêtre, France \\
guillaume.tochon@lrde.epita.fr}
\and
\IEEEauthorblockN{Mauro Dalla Mura}
\IEEEauthorblockA{\textit{Univ. Grenoble Alpes, CNRS, Grenoble INP, GIPSA-lab} \\
\textit{Institut Universitaire de France}\\
Grenoble, France \\
mauro.dalla-mura@gipsa-lab.grenoble-inp.fr}
\and
\IEEEauthorblockN{Abdeldjalil Aissa El Bey}
\IEEEauthorblockA{\textit{Lab-STICC} \\
\textit{IMT Atlantique}\\
Brest, France \\
abdeldjalil.aissaelbey@imt-atlantique.fr}
}

\maketitle

\begin{abstract}
In the context of an increasing popularity of data-driven models to represent dynamical systems, many machine learning-based implementations of the Koopman operator have recently been proposed. However, the vast majority of those works are limited to deterministic predictions, while the knowledge of uncertainty is critical in fields like meteorology and climatology. In this work, we investigate the training of ensembles of models to produce stochastic outputs. We show through experiments on real remote sensing image time series that ensembles of independently trained models are highly overconfident and that using a training criterion that explicitly encourages the members to produce predictions with high inter-model variances greatly improves the uncertainty quantification of the ensembles.
\end{abstract}


\section{Introduction}

With the simultaneously growing availability of observed geophysical data and advancement in machine learning methods, recent data-driven models have shown impressive performance in accurately forecasting physical dynamical systems~\cite{lam2022graphcast}. Despite their performance, these models are harder to interpret than traditional methods, which means that it is more difficult to trust their predictions. One way to partially circumvent this flaw is to design models that can produce probability distributions of predictions instead of outputting a single prediction. 
Such models are then able to quantify the uncertainty, as the variance of a model's prediction can be seen as a measure of confidence.
In this regard, one must be aware of the different sources of uncertainty that are to be identified. In the machine learning community, the uncertainty is usually decomposed into two categories: aleatoric uncertainty is the uncertainty that comes from the training data, e.g. due to noise and/or scarce sampling, while epistemic uncertainty denotes everything that comes from the model, e.g. the incapacity to fit the training data due to a lack of expressivity. We refer to~\cite{haynes2023creating} for an extensive discussion on the sources of uncertainty.

Our work is based on the Koopman operator theory~\cite{koopman1931hamiltonian}, which states that any nonlinear dynamical system can be described by a linear operator acting on the set of its measurement functions. While the Koopman operator is infinite dimensional in practice, many methods have been proposed to find approximate finite-dimensional representations, with applications to e.g. fluid dynamics and epidemiology~\cite{brunton2021modern}.

We are most interested here in finite-dimensional representations based on neural auto-encoders~\cite{lusch2018deep,otto2019linearly,frion2023leveraging}. These methods aim to learn a mapping from the input space of a dynamical system to a finite set of observation functions which are stable under the action of the Koopman operator, and vice versa. Using the machine learning vocabulary, the obtained Koopman invariant subspace is defined by the latent space of a learnt auto-encoder. 
Another learnt component (generally a matrix $\mathbf{K}$) then governs the evolution of the latent state through time.
However, unlike for DMD-based methods, most of the existing works on Koopman autoencoders consider only deterministic models, which are unable to provide uncertainty estimates. Here, we are interested in finding simple ways to adapt these models to stochastic contexts. While there are many such ways in deep learning~\cite{gal2016dropout, malinin2018predictive, blundell2015weight}, we focus on deep ensembles, which are computationally intensive at training and inference but require no architectural change to a deterministic model and still outperform bayesian methods in some cases~\cite{lakshminarayanan2017simple}. We show that in our case the usual way of training the members of an ensemble independently from each other leads to a highly overconfident ensemble. In order to alleviate this issue, we propose to train the members jointly with a loss function that encourages them to produce more diversified predictions. 

The remainder of this paper is organised as follows: in section~\ref{sec:related_works}, we review previous contributions on Koopman autoencoder models and on training neural networks for uncertainty quantification, especially with ensembles of models. We then introduce new methods for training ensembles of Koopman autoencoders in section~\ref{sec:methods}, and use them for forecasting time series of multispectral satellite images in section~\ref{sec:experiments}.

\section{Related works}
\label{sec:related_works}

\subsection{Data-driven implementations of the Koopman operator}
\label{sec:KAE}

Although the Koopman operator theory~\cite{koopman1931hamiltonian} dates back to the 1930s, it has known renewed interest when data-driven models related to this theory were introduced in the past two decades, notably dynamic mode decomposition (DMD)~\cite{DMD} and its extensions, which are reviewed in~\cite{brunton2021modern}.
We will focus more specifically on recent methods~\cite{lusch2018deep, otto2019linearly, frion2023leveraging} which consist in using a neural autoencoder to learn a finite set of observation functions on which the general evolution of a dynamical system can be described linearly. Most of the approaches in this line jointly learn an encoder $\phi : \mathbb{R}^n \to \mathbb{R}^d$, a decoder $\psi : \mathbb{R}^d \to \mathbb{R}^n$ and a matrix $\mathbf{K} \in \mathbb{R}^{d \times d}$ such that the advancement of an initial state $\mathbf{x}_0 \in \mathbb{R}^n$ by a time $\tau$ through the modelled dynamical system can be (approximately) written as
\begin{equation}
\label{eq:model}
    \mathbf{x}_\tau = \psi(\mathbf{K}^\tau\phi(\mathbf{x}_0))
\end{equation}
While this equation is often understood as a discrete model, where $\tau$ may only be a positive integer, a continuous formulation has been discussed in~\cite{otto2019linearly} and later implemented in~\cite{frion2023neural}.

\subsection{Uncertainty quantification for neural networks}

Uncertainty quantification for neural networks is a field that has recently gained a lot of interest. 
We refer the interested reader to~\cite{gawlikowski2021survey} for an in-depth survey of uncertainty quantification for neural networks, and to~\cite{haynes2023creating} for a very recent review that focuses on environmental-science applications.
One of the most popular approaches to evaluate the quality of uncertainty estimates is the continuous ranked probability score (CRPS). First introduced in~\cite{matheson1976scoring}, the CRPS is defined as
\begin{equation}
    \text{CRPS}(F,y_{true}) = \int_{-\infty}^\infty [F(y) - \mathds{1}_{y \geq y_{true}}]^2 dy,
    \end{equation}
where $F$ is the cumulated distribution function of the output distribution, $y_{true}$ is the groundtruth and $\mathds{1}_{y \geq y_{true}}$ is the Heaviside function, taking value $1$ for $y \geq y_{true}$ and $0$ otherwise. When the output distribution is a set of $M$ ensemble members $(y_j)_{j=1}^M$, the CRPS can be reframed~\cite{gneiting2007strictly} as
\begin{equation}
\label{eq:crps_mae}
    \text{CRPS} = \frac{1}{M}\sum_{j=1}^M |y_{true}-y_i| - \frac{1}{2}\frac{1}{M^2}\sum_{j=1}^M\sum_{k=1}^M |y_j - y_k|.
\end{equation}
The first term is the mean absolute error (MAE) of the prediction and the second term is the halved mean absolute pairwise difference between ensemble members. Thus, the CRPS of a determinstic prediction is simply its MAE. The CRPS is being increasingly used as a loss function, e.g. in~\cite{Baran:2021} with a parametric model for which the output is the mean and variance of a gaussian distribution. When the groundtruth is multivariate, the above expressions can simply be summed over all variables (hence ignoring correlations).

Let us now discuss more specifically the works on ensembles of neural networks.
These ensembles generally consist in several instances of a same model, which can differ by various factors such as their initial parameterization or the set of data that they have been trained on. While the motivation of training ensembles of models in machine learning was at first to boost the performance by averaging the predictions of the ensemble members~\cite{dietterich2000ensemble}, it has been later noticed~\cite{lakshminarayanan2017simple} that the variance of the predictions of an ensemble's members can also be used as a natural estimation of the uncertainty of the ensemble. However, in contrast to the methods that we propose in section~\ref{sec:methods}, \cite{lakshminarayanan2017simple} identified the independence of the trained members of the ensemble as a key element in training a deep ensemble, and the vast majority of subsequent works followed this principle, with the notable exception of~\cite{jain2020maximizing}.

\section{Proposed methods}
\label{sec:methods}

\subsection{Introduction of a variance-promoting loss term}

We train several instances of the model from~\cite{frion2023neural}, which is a classical Koopman autoencoder with 3 components, $\phi$, $\psi$ and $\mathbf{K}$ as described in subsection~\ref{sec:KAE}.

In the single-instance version of this model, we denote by $\theta$ the set of all trainable parameters (including the coefficients of $\mathbf{K}$ and the trainable parameters of $(\phi, \psi)$). Suppose that we are working with a $n$-dimensional dynamical system and that our training dataset $(\mathbf{x}_{i,t})_{1\leq i \leq N, 0\leq t \leq T}$ is composed of $N$ time series of length $T+1$ resulting from the dynamical system of interest. Note that $\mathbf{x}_{i,t}$ is a $n$-dimensional vector. In what follows, we may drop the index $i$ to designate any of these time series. The loss function is composed of the terms:
\begin{align}
\label{eq:pred_loss}
    &L_{pred}(\theta) = 
    \sum_{1\leq i \leq N}
    \sum_{1 \leq \tau \leq T}
        ||\mathbf{x}_{i,\tau} - \psi(\mathbf{K}^{\tau}\phi(\mathbf{x}_{i,0}))||^2 \\
    &\label{eq:ae_loss}L_{ae}(\theta) = 
        \sum_{1 \leq i \leq N}
        \sum_{0 \leq t \leq T}
        ||\mathbf{x}_{i,t} - \psi(\phi(\mathbf{x}_{i,t}))||^2\\
    &\label{eq:lin_loss}
    L_{lin}(\theta) = 
        \sum_{1 \leq i \leq N}
        \sum_{1 \leq \tau \leq T}
        ||\phi(\mathbf{x}_{i,\tau}) - \mathbf{K}^{\tau}\phi(\mathbf{x}_{i,0})||^2\\
    &\label{eq:orth_loss}L_{orth}(\theta) = L_{orth}(\mathbf{K}) = ||\mathbf{KK}^T - \mathbf{I}||_{F}^2
\end{align}
where $||.||_F$ denotes the Frobenius norm.
The terms $L_{pred}$, $L_{ae}$, $L_{lin}$ and $L_{orth}$ are respectively the prediction loss, auto-encoding loss, linearity loss and orthogonality loss. We refer to~\cite{frion2023neural} for further interpretation. They can all be weighted equally except for the orthogonality loss for which a suitable weight $\alpha$ has to be found, resulting in the global loss function
\begin{equation}
\label{eq:4_terms}
    L(\theta) = L_{pred}(\theta) + L_{ae}(\theta) + L_{lin}(\theta) + \alpha L_{orth}(\theta).
\end{equation}
Let us now suppose that we are training an ensemble of $M$ instances of this model. We then denote the parameters of these instances as $\Theta = (\theta_1, ..., \theta_M)$. The instances can be trained in parallel by defining a global loss function which is simply the sum of the loss functions for each of the $M$ instances:
\begin{equation}
\label{eq:independent_training}
    \mathcal{L}_{independent}(\Theta) = \frac{1}{M} \sum_j L(\theta_j).
\end{equation}
In this equation and in the following ones, all sums are defined on index $j$ from $1$ to $M$. 
Using this loss function is equivalent to training the $M$ members of the ensemble sequentially and independently. As we will show experimentally, this may lead to a low diversity of the members, since the instances tend to all capture similar features in the data, which is undesirable in ensemble learning.
Given an input state $\mathbf{x}_0 \in \mathbb{R}^n$, the outputs of the members, obtained by equation~\eqref{eq:model}, are denoted as $\hat{\mathbf{x}}_{t,j}$ with $0 \leq t \leq T$ denoting time and $1 \leq j \leq M$ denoting the member. The mean prediction of the members is defined as
\begin{equation}
\label{eq:empirical_mean}
    \hat{\mathbf{x}}_t = \frac{1}{M} \sum_j \hat{\mathbf{x}}_{t,j}.
\end{equation}

For the uncertainty quantification to be accurate, we would like the empirical variance of these predictions to be close to the squared error between the mean prediction and the groundtruth $\mathbf{x}_t$: see~\cite{haynes2023creating} for reference. Thus, we seek
\begin{equation}
\label{eq:spread-skill}
    \frac{1}{M-1}\sum_j||\hat{\mathbf{x}}_{t,j} - \hat{\mathbf{x}}_t||^2 \approx ||\hat{\mathbf{x}}_t - \mathbf{x}_t||^2.
\end{equation}
In practice, since the empirical variance is too low when training an ensemble with equation~\eqref{eq:independent_training}, we introduce a variance-promoting loss term, which takes into account all members:
\begin{equation}
\label{eq:var_loss}
    L_{var}(\Theta) = - \frac{1}{M}\sum_j||\hat{\mathbf{x}}_{t,j} - \hat{\mathbf{x}}_t||^2.
\end{equation}
We write this term with simplified notations, yet the full loss term includes sums over $i$ and $t$ like in equations~\eqref{eq:pred_loss} to~\eqref{eq:lin_loss}. Using the new loss term from equation~\eqref{eq:var_loss}, one can now introduce a new loss function for training ensembles:
\begin{equation}
\label{eq:var_training}
    \mathcal{L}_{var,\lambda}(\Theta) =  \mathcal{L}_{independent}(\Theta) + \lambda L_{var}(\Theta),
\end{equation}
where the case $\lambda=0$ corresponds to equation~\eqref{eq:independent_training}. This loss promotes the sum of variances over each of the $N$ variables of the state individually, i.e. the total variance of the predictions. 

\subsection{Analysis of the choice of $\lambda$}

It is a standard and easy to prove result in statistics that
\begin{equation}
\label{eq:Koenig-Huygens}
    \frac{1}{M}\sum_j ||\hat{\mathbf{x}}_{t,j} - \mathbf{x}_t||^2 = \frac{1}{M}\sum_j||\hat{\mathbf{x}}_{t,j} - \hat{\mathbf{x}}_t||^2 + ||\hat{\mathbf{x}}_t - \mathbf{x}_t||^2
\end{equation}
for any value $\mathbf{x}_t$ and predictions $\hat{\mathbf{x}}_{t,j}$. From this, we obtain
\begin{multline}
\label{eq:pred+var}
\frac{1}{M} \sum_j ||\hat{\mathbf{x}}_{t,j} - \mathbf{x}_t||^2 - \frac{\lambda}{M}\sum_j||\hat{\mathbf{x}}_{t,j} - \hat{\mathbf{x}}_t||^2 = \\ 
\frac{1-\lambda}{M}\sum_j||\hat{\mathbf{x}}_{t,j} - \hat{\mathbf{x}}_t||^2 + ||\hat{\mathbf{x}}_t - \mathbf{x}_t||^2
\end{multline}

for a given $\lambda$. If $\lambda \leq 1$, then this expression is trivially positive. However, if $\lambda > 1$, then it is not negatively bounded. Indeed, since $\mathbf{x}_t$ is a constant vector, one can simply choose arbitrarily large member predictions $\hat{\mathbf{x}}_{t,j}$ satisfying $\hat{\mathbf{x}}_t = 0$ through equation~\eqref{eq:empirical_mean} (e.g. the members can be arranged in opposite pairs) in order for the expression~\eqref{eq:pred+var} to become arbitrarily low. 
As this analysis remains true for any $\mathbf{x}_0$ and $t$, the prediction loss term~\eqref{eq:pred_loss} can counterbalance the variance loss term~\eqref{eq:var_loss} as long as $0 \leq \lambda \leq 1$ in equation~\eqref{eq:var_training}. If $\lambda > 1$ then the training procedure will diverge with an inter-model variance growing to infinity. 
Therefore, the hyperparameter $\lambda$ for training the ensemble should be chosen between 0 and 1, and the variance of the predictions grows with the value of $\lambda$, as we will show in our experiments. This simple interpretation of $\lambda$ motivates the use of a biased estimator of the variance in equation~\eqref{eq:var_loss}: with an unbiased estimator, the acceptable range for $\lambda$ would depend on the number $M$ of members.

To the best of our knowledge, the introduction of this loss term for training an ensemble of neural networks is a novelty. The closest contribution that we identified was in~\cite{jain2020maximizing}, which introduced loss terms similar to~\eqref{eq:var_loss} with a notable difference being that the authors work with models that have bounded outputs, so that the parameter $\lambda$ can be set arbitrarily without fear of obtaining variances that diverge to infinity. The choice of unbounded models trades more flexibility in the individual members with the constraints on $\lambda$ that we described above.

\subsection{Using a loss term inspired by the CRPS}

An alternative to the loss function~\eqref{eq:var_training} is to use the CRPS for training. While easy to compute for small ensembles, the formulation of equation~\eqref{eq:crps_mae} gets very costly to compute as the number $M$ of members gets higher because of the pairwise differences in the second term. Therefore, we propose to replace this term by the mean absolute error between the individual predictions and the mean prediction. Using the same notations as in equation~\eqref{eq:var_loss}, we introduce
\begin{equation}
\label{eq:lin_diversity_loss}
    L_{abs}(\Theta) = -\frac{1}{2}\frac{1}{M}\sum_j|\hat{\mathbf{x}}_{t,j} - \hat{\mathbf{x}}_t|.
\end{equation}
One can easily prove that
\begin{equation}
\begin{split}
    \frac{1}{M}\sum_j|\hat{\mathbf{x}}_{t,j} - \hat{\mathbf{x}}_t| 
    & \leq \frac{1}{M^2}\sum_{j=1}^M\sum_{k=1}^M |\hat{\mathbf{x}}_{t,j} - \hat{\mathbf{x}}_{t,k}| \\
    & \leq \frac{2}{M}\sum_j|\hat{\mathbf{x}}_{t,j} - \hat{\mathbf{x}}_t|,
\end{split}
\end{equation}
so that~\eqref{eq:lin_diversity_loss} can be seen as a proxy to the second term of the CRPS formulation in~\eqref{eq:crps_mae}, while the first term is analogous to the sum of the prediction losses~\eqref{eq:pred_loss} using the $\mathcal{L}_1$ distance instead of the squared $\mathcal{L}_2$ distance. This motivates the introduction of a new diversity-promoting loss function:
\begin{equation}
\label{eq:proxy_crps_training}
    \mathcal{L}_{CRPS}(\Theta) = \sum_j L_1(\theta_j) + \lambda L_{abs}(\Theta),
\end{equation}
where $\lambda$ is usually set to 1 and $L_1(\theta)$ is a variant of equation~\eqref{eq:4_terms} where all squared $\mathcal{L}_2$ distances are replaced by $\mathcal{L}_1$ distances for the purpose of consistency/homogeneity between all loss terms. Note that this loss is not the true CRPS but an approximation of it, with auxiliary terms.

\section{Experiments}
\label{sec:experiments}

In this section, we present our experiments on datasets originally introduced in~\cite{frion2023learning}, and consisting of time series of Sentinel-2 multispectral satellite images over two spatial areas: the forest of Fontainebleau and the forest of Orléans in France. The datasets and codes are available at \url{https://github.com/anthony-frion/Sentinel2TS}. The time series for the two areas are available in two versions. The original versions contain raw images with no pre-processing other than the classical
(level-2A) atmospheric correction, yet the time series are incomplete as only the images that are not corrupted by the presence of clouds are retained. Since training from irregularly-sampled time series is very challenging, the second versions interpolate these available images in time through Cressman interpolation, resulting in partly synthetic but regularly-sampled time series. Although Koopman autoencoders are able to handle irregularly-sampled time series~\cite{frion2023neural}, we choose to train on the regular version of the Fontainebleau data in order to keep the training procedure simple. We test the trained ensembles on two tasks: 1) extrapolating on the training Fontainebleau area to times unseen during training 2) predicting from an initial time on the test Orléans area. Thus, task 1) is used to test temporal extrapolation while task 2) tests the ability to transfer the knowledge to a new area with a distribution shift.

\begin{figure}
    \centering
    \includegraphics[width=8.5cm]{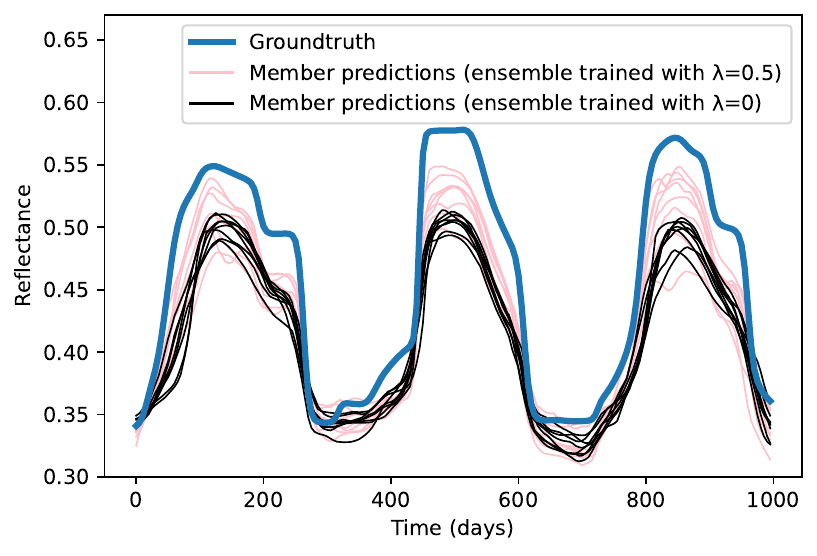}
    \caption{Forecasting from time 0 by two ensembles for the reflectance of the B7 band (in near infrared) for a Fontainebleau pixel. Here, both ensembles are biased, but the ensemble trained with a variance-promoting loss term ($\lambda=0.5$) yields a higher inter-member variance, and hence a better uncertainty estimate, than the ensemble of independently trained models ($\lambda=0$).}
    \label{fig:overconfident_pred}
\end{figure}

We first motivated the introduction of our customized ensemble training loss~\eqref{eq:var_training} by making the simple observation that independently trained members from an ensemble of Koopman autoencoders tend to learn very similar dynamics. We show on figure~\ref{fig:overconfident_pred} a typical example for an ensemble trained with loss~\eqref{eq:independent_training} (i.e. $\lambda=0$ in loss~\eqref{eq:var_training}), where all 8 instances make similar predictions from an initial observation belonging to the training area. Here, the member predictions are all much closer to each other than any of them is to the groundtruth. Although this is an illustrative example with a relatively high forecasting error, it is symptomatic of a very overconfident ensemble. We also show the predictions of an ensemble trained in the same conditions but with $\lambda=0.5$ in loss~\eqref{eq:var_training}: although this ensemble is biased too, its higher variance makes it less overconfident, and thus better in this case. 

In order to promote the diversity of the members, we now train ensembles with the loss function~\eqref{eq:var_training} with different values of $\lambda$, remembering that $\lambda=0$ corresponds to training the models independently from one another, as classically done in the literature. As expected, the variance of the predictions increases with $\lambda$.
We also train an ensemble with the loss function~\eqref{eq:proxy_crps_training}. We evaluate the quality of the uncertainty estimates with 2 methods: the CRPS and the spread-skill plot.

\begin{figure}
    \centering
    \includegraphics[width=8.5cm]{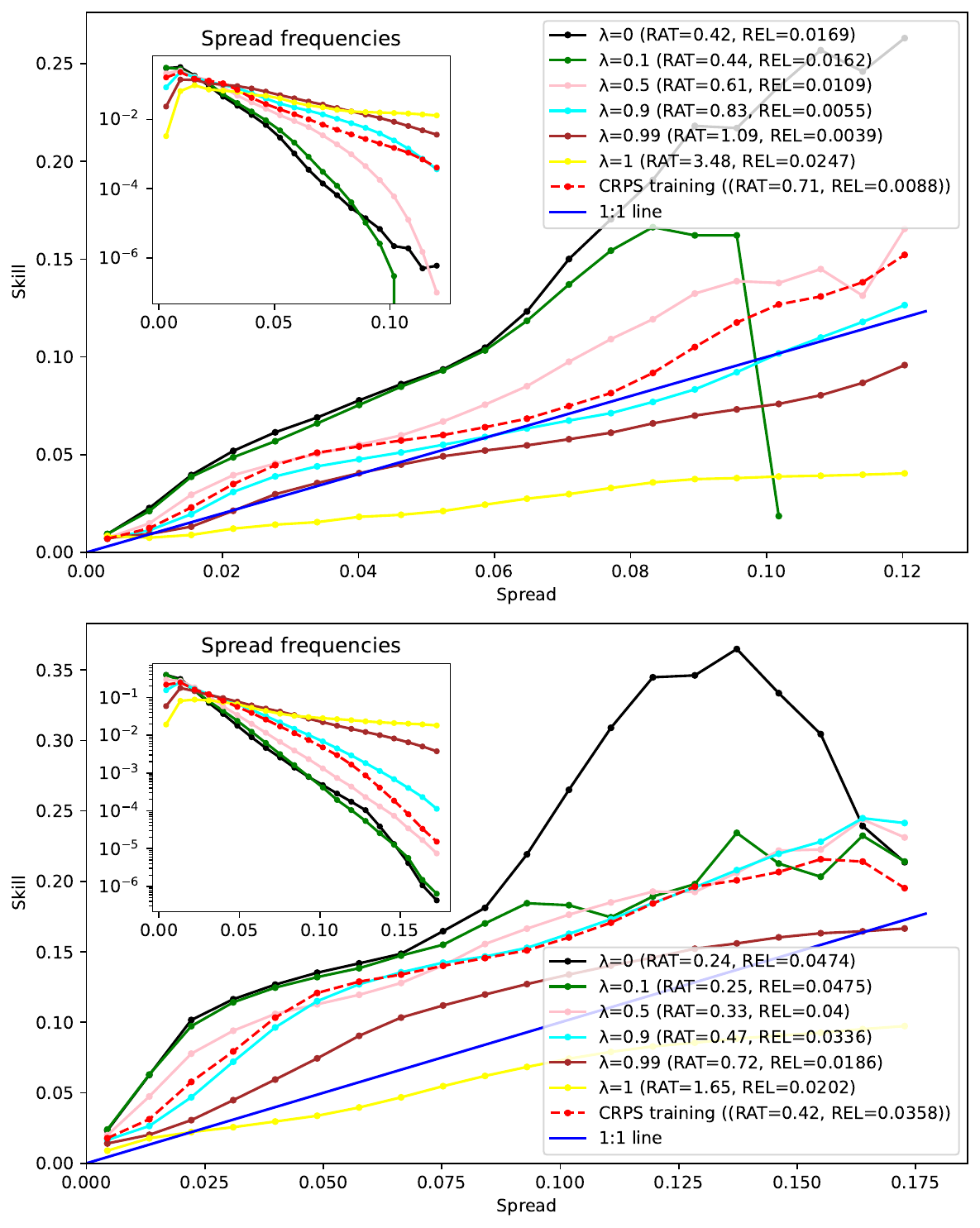}
    \caption{Spread-skill plots for two different datasets. Top: spread-skill plot of extrapolation on the training Fontainebleau area. Bottom: spread-skill plot of predictions from time 0 on the test Orléans area.}
    \label{fig:spread_skill}
\end{figure}

We show in figure~\ref{fig:spread_skill} the spread-skill plots for the two identified tasks. The idea of these plots is to represent the skill as a function of the spread. The skill is defined as the root mean squared error between the average prediction of the ensemble (obtained by equation~\eqref{eq:empirical_mean}) and the groundtruth. The spread is defined as the standard deviation of the predictions of the members.
The spread and skill are the square roots of the left and right members of equation~\eqref{eq:spread-skill}: one would like them to be approximately equal.
In practice, we consider a set of ensemble predictions and groundtruth, where all spectral bands and prediction time spans are separated, resulting in univariate values. We compute a 20-bin histogram of this set according to the spread. Then, for each bin, we compute the mean skill of the predictions and the number of points inside the bin. On the main plot, each point corresponds to one of the bins, and we would like the points to stay close to the 1:1 line, which corresponds to equation~\eqref{eq:spread-skill}. If the plot is above the 1:1 line, it means that the ensemble tends to underestimate its errors, hence it is overconfident. On the contrary, a plot below the 1:1 line means that the ensemble is underconfident. The inner plots show the frequencies associated to each bin.

A spread-skill plot can be summarized by two metrics: the SSREL is the sum of the absolute distances to the 1:1 line over the bins, weighted by the bin frequencies. It is positive and its ideal value is zero. The SSRAT measures the spread-skill ratio globally, and is unaffected by the binning process. It is positive and unitless, and a value smaller or greater than $1$ respectively characterize an overconfident or an underconfident model.
We refer to~\cite{haynes2023creating} for more extensive discussions on these notions.

Several conclusions can be drawn from figure~\ref{fig:spread_skill}. First, the ensemble with independently trained models (corresponding to $\lambda=0$) is highly overconfident, which quantitatively confirms the intuition gained from figure~\ref{fig:overconfident_pred}. Then, one can clearly see that the ensembles get less confident as the value of $\lambda$ increases. The case of $\lambda=1$ is a limit case, and results in the only model that is severely underconfident on both training and test areas. The value $\lambda=0.99$ yields the best spread-skill ratios, and the model trained with a proxy to the CRPS lies in between $\lambda=0.5$ and $\lambda=0.9$.

\begin{figure}
    \centering
    \includegraphics[width=8.5cm]{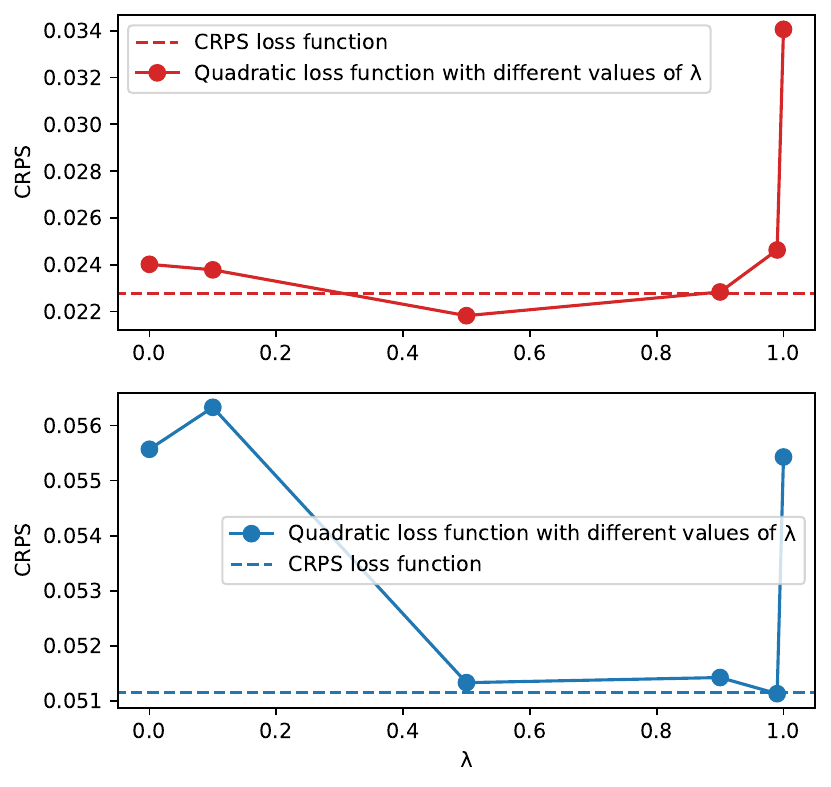}
    \caption{CRPS of ensembles of Koopman autoencoders according to the weight $\lambda$ of their variance-promoting loss term during training. Top: extrapolation on training Fontainebleau area. Bottom: transfer to test Orléans area. The represented values of $\lambda$ are $0, 0.1, 0.5, 0.9, 0.99, 1$.}
    \label{fig:crps}
\end{figure}

Finally, we show in figure~\ref{fig:crps} the CRPS of the ensembles as a function of the value of $\lambda$ used in their training function~\eqref{eq:var_training}. Note that lower values are better for the CRPS metric. 
Again, one can see that a well-chosen value of $\lambda$ can significantly improve the performance compared to an ensemble of independently trained members ($\lambda=0)$. 
The values $\lambda=0.5$ and $\lambda=0.9$ seem to be good compromises between the CRPS on the two tasks, while the model trained with a loss function similar to the CRPS also performs well on both.

\section{Conclusion}

In this work, after noticing that ensembles of Koopman autoencoders tend to be very overconfident when their members are trained independently, we introduced a variance-promoting loss term which encourages the members of an ensemble to produce more diverse forecasts. We studied, both analytically and empirically, the influence of this term on the trained ensembles according to its weight relatively to the other loss terms. We found that, according to several metrics, the quality of the uncertainties produced by the ensembles improves as the weight of the variance-promoting loss term gets closer to its theoretical limit of $1$. In future works, we will try using this loss term in conjunction with other uncertainty quantification methods, e.g. Monte Carlo dropout and ensemble predictions with a single model. We also intend to further study the specificities of uncertainty quantification for long-term forecasting tasks, and for Koopman autoencoders in particular.

\bibliographystyle{IEEEtran}
\bibliography{references.bib}
\end{document}